\documentclass[11pt,a4paper]{article}
\usepackage[hyperref]{acl2018}
\usepackage{times}
\usepackage{latexsym}
\usepackage{CJK}
\usepackage{url}
\usepackage{natbib}
\usepackage{amsmath}
\usepackage{amsfonts,amssymb}
\usepackage{graphicx}
\usepackage{algorithm}
\usepackage{algorithmicx}
\usepackage{algpseudocode}
\newcommand*{\affaddr}[1]{#1} 
\newcommand*{\affmark}[1][*]{\textsuperscript{#1}}
\newcommand*{\email}[1]{\texttt{#1}}
\aclfinalcopy 
\def\aclpaperid{1281} 

\title{Unpaired Sentiment-to-Sentiment Translation: A Cycled\\ Reinforcement Learning Approach}

\author{Jingjing Xu\affmark[1]\footnotemark[1],  \ Xu Sun\affmark[1]\Thanks{\ Equal Contribution.}, \ Qi Zeng\affmark[1], \ Xuancheng Ren\affmark[1], \\\textbf{Xiaodong Zhang\affmark[1], \ Houfeng Wang\affmark[1], \ Wenjie Li\affmark[2]}\\
\affaddr{\affmark[1]MOE Key Lab of Computational Linguistics, School of EECS, Peking University}\\
\affaddr{\affmark[2]Department of Computing, Hong Kong Polytechnic University}\\
\email{\{jingjingxu,xusun,pkuzengqi,renxc,zxdcs,wanghf\}@pku.edu.cn}\\
\email{cswjli@comp.polyu.edu.hk}\\
}

\date{}

\begin{document}
\maketitle

\begin{abstract}

The goal of sentiment-to-sentiment ``translation'' is to change the underlying sentiment of a sentence while keeping its content. The main challenge is the lack of parallel data. To solve this problem, we propose a cycled reinforcement learning method that enables training on unpaired data by collaboration between a neutralization module and an emotionalization module. We evaluate our approach on two review datasets, Yelp and Amazon. Experimental results show that our approach significantly outperforms the state-of-the-art systems. Especially, the proposed method substantially improves the content preservation performance. The BLEU score is improved from 1.64 to 22.46 and from 0.56 to 14.06 on the two datasets, respectively.\footnote{The released code can be found in https://github.com/lancopku/unpaired-sentiment-translation}
 
\end{abstract}

\section{Introduction}

Sentiment-to-sentiment ``translation'' requires the system to change the underlying sentiment of a sentence while preserving its non-emotional semantic content as much as possible. It can be regarded as a special style transfer task that is important in Natural Language Processing (NLP)~\citep{Hu17controlled,Shen17crossalign,DBLP:journals/corr/abs-1711-06861}. It has broad applications, including review sentiment transformation, news rewriting, etc. Yet the lack of parallel training data poses a great obstacle to a satisfactory performance. 

Recently, several related studies for language style transfer~\citep{Hu17controlled,Shen17crossalign} have been proposed. However, when applied to the sentiment-to-sentiment ``translation'' task, most existing studies only change the underlying sentiment and fail in keeping the semantic content. For example, given ``The food is delicious" as the source input, the model generates ``What a bad movie" as the output. Although the sentiment is successfully transformed from positive to negative, the output text focuses on a different topic. The reason is that these methods attempt to implicitly separate the emotional information from the semantic information in the same dense hidden vector, where all information is mixed together in an uninterpretable way. Due to the lack of supervised parallel data, it is hard to only modify the underlying sentiment without any loss of the non-emotional semantic information.

To tackle the problem of lacking parallel data, we propose a cycled reinforcement learning approach that contains two parts: a neutralization module and an emotionalization module. The neutralization module is responsible for extracting non-emotional semantic information by explicitly filtering out emotional words. The advantage is that only emotional words are removed, which does not affect the preservation of non-emotional words. The emotionalization module is responsible for adding sentiment to the neutralized semantic content for sentiment-to-sentiment translation.

In cycled training, given an emotional sentence with sentiment $s$, we first neutralize it to the non-emotional semantic content, and then force the emotionalization module to reconstruct the original sentence by adding the sentiment $s$. Therefore, the emotionalization module is taught to add sentiment to the semantic context in a supervised way. By adding opposite sentiment, we can achieve the goal of sentiment-to-sentiment translation. Because of the discrete choice of neutral words, the gradient is no longer differentiable over the neutralization module. Thus, we use policy gradient, one of the reinforcement learning methods, to reward the output of the neutralization module based on the feedback from the emotionalization module. We add different sentiment to the semantic content and use the quality of the generated text as reward. The quality is evaluated by two useful metrics: one for identifying whether the generated text matches the target sentiment; one for evaluating the content preservation performance. The reward guides the neutralization module to better identify non-emotional words. In return, the improved neutralization module further enhances the emotionalization module.

Our contributions are concluded as follows:

\begin{itemize}
\item For sentiment-to-sentiment translation, we propose a cycled reinforcement learning approach. It enables training with unpaired data, in which only reviews and sentiment labels are available.

\item Our approach tackles the bottleneck of keeping semantic information by explicitly separating sentiment information from semantic content. 

\item Experimental results show that our approach significantly outperforms the state-of-the-art systems, especially in content preservation.
\end{itemize}

\section{Related Work}

Style transfer in computer vision has been studied \citep{DBLP:conf/eccv/JohnsonAF16,DBLP:conf/cvpr/GatysEB16,DBLP:journals/tog/LiaoYYHK17,DBLP:conf/ijcai/li,DBLP:conf/iccv/ZhuPIE17}. The main idea is to learn the mapping between two image domains by capturing shared representations or correspondences of higher-level structures.

There have been some studies on unpaired language style transfer recently.  \newcite{Hu17controlled} propose a new neural generative model that combines variational auto-encoders (VAEs) and holistic attribute discriminators for effective imposition of style semantic structures. \newcite{DBLP:journals/corr/abs-1711-06861} propose to use an adversarial network to make sure that the input content does not have style information. \newcite{Shen17crossalign} focus on separating the underlying content from style information. They learn an encoder that maps the original sentence to style-independent content and a style-dependent decoder for rendering. However, their evaluations only consider the transferred style accuracy. We argue that content preservation is also an indispensable evaluation metric. However, when applied to the sentiment-to-sentiment translation task, the previously mentioned models share the same problem. They have the poor preservation of non-emotional semantic content.

In this paper, we propose a cycled reinforcement learning method to improve sentiment-to-sentiment translation in the absence of parallel data. The key idea is to build supervised training pairs by reconstructing the original sentence.  A related study is ``back reconstruction'' in machine translation~\cite{DBLP:conf/nips/HeXQWYLM16,DBLP:conf/aaai/TuLSLL17}. They couple two inverse tasks: one is for translating a sentence in language $A$ to a sentence in language $B$; the other is for translating a sentence in language $B$ to a sentence in language $A$. Different from the previous work, we do not introduce the inverse task, but use collaboration between the neutralization module and the emotionalization module.

Sentiment analysis is also related to our work~\cite{DBLP:conf/icml/SocherLNM11, DBLP:conf/semeval/PontikiGPMA15,DBLP:conf/semeval/RosenthalFN17,DBLP:journals/eswa/ChenXHW17,DBLP:conf/ijcnlp/MaLZWS17,shumingmaijcai}. The task usually involves detecting whether a piece of text expresses positive, negative, or neutral sentiment. The sentiment can be general or about a specific topic.

\section{Cycled Reinforcement Learning for Unpaired Sentiment-to-Sentiment Translation}
In this section, we introduce our proposed method. An overview is presented in Section~\ref{overview}. The details of the neutralization module and the emotionalization module are shown in Section~\ref{Neutralization} and Section~\ref{Emotionalization}. The cycled reinforcement learning mechanism is introduced in Section~\ref{dual}.

\subsection{Overview}\label{overview}
\begin{figure}[t] 
\centering
\includegraphics[width = 0.9\linewidth]{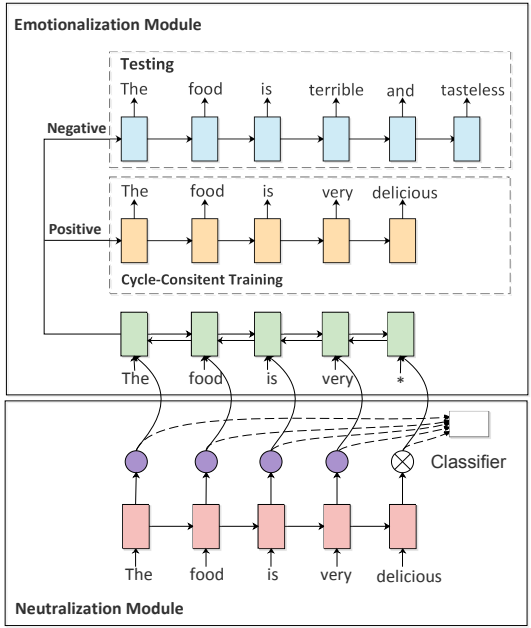}
\caption{An illustration of the two modules. Lower: The neutralization module removes emotional words and extracts non-emotional semantic information. Upper: The emotionalization module adds sentiment to the semantic content. The proposed self-attention based sentiment classifier is used to guide the pre-training.}
\label{structure} 
\end{figure}

The proposed approach contains two modules: a neutralization module and an emotionalization module, as shown in Figure~\ref{structure}. The neutralization module first extracts non-emotional semantic content, and then the emotionalization module attaches sentiment to the semantic content. Two modules are trained by the proposed cycled reinforcement learning method.  The proposed method requires the two modules to have initial learning ability. Therefore, we propose a novel pre-training method, which uses a self-attention based sentiment classifier (SASC). A sketch of cycled reinforcement learning is shown in Algorithm~\ref{code}. The details are introduced as follows. 

\begin{algorithm}[t]
\caption{The cycled reinforcement learning method for training the neutralization module $N_{\theta}$ and the emotionalization module $E_{\phi}$.}
\label{code} 
\small
  \begin{algorithmic}[1]
  
      \State Initialize the neutralization module $N_{\theta}$, the emotionalization module $E_{\phi}$ with random weights $\theta$, $\phi$ 
      \State Pre-train $N_{\theta}$ using MLE based on Eq. \ref{newi} 
      \State Pre-train $E_{\phi}$ using MLE based on Eq. \ref{em}
    \For{each iteration $i=1,2,..., M$}  
    	\State Sample a sequence $\boldsymbol{x}$ with sentiment $s$ from $X$
        \State Generate a neutralized sequence $\hat{\boldsymbol{x}}$ based on $N_{\theta}$ 
        \State Given $\hat{\boldsymbol{x}}$ and $s$, generate an output based on $E_{\phi}$
         \State Compute the gradient of $E_{\phi}$ based on Eq. \ref{dual-learning-2}
        \State Compute the reward $R_1$ based on Eq.~\ref{reward} 
        \State $\bar{s}$ = the opposite sentiment
         \State Given $\hat{\boldsymbol{x}}$ and $\bar{s}$, generate an output based on $E_{\phi}$
          \State Compute the reward $R_2$ based on Eq.~\ref{reward}
        \State Compute the combined reward $R_{c}$ based on Eq.~\ref{combined_reward}
        \State Compute the gradient of $N_{\theta}$ based on Eq. \ref{dual-learning-1}
        \State Update model parameters $\theta$, $\phi$
    \EndFor  
     
  \end{algorithmic} 
  
\end{algorithm}  

\subsection{Neutralization Module}\label{Neutralization}

The neutralization module $N_\theta$ is used for explicitly filtering out emotional information. In this paper, we consider this process as an extraction problem. The neutralization module first identifies non-emotional words and then feeds them into the emotionalization module. We use a single Long-short Term Memory Network (LSTM) to generate the probability of being neutral or being polar for every word in a sentence.  Given an emotional input sequence $\boldsymbol{x} = (x_1, x_2, \ldots, x_T)$ of $T$ words from $\Gamma$, the vocabulary of words, this module is responsible for producing a neutralized sequence. 

Since cycled reinforcement learning requires the modules with initial learning ability, we propose a novel pre-training method to teach the neutralization module to identify non-emotional words. We construct  a self-attention based sentiment classifier and use the learned attention weight as the supervisory signal. The motivation comes from the fact that, in a well-trained sentiment classification model, the attention weight reflects the sentiment contribution of each word to some extent. Emotional words tend to get higher attention weights while neutral words usually get lower weights. The details of sentiment classifier are described as follows. 

Given an input sequence $\boldsymbol{x}$, a sentiment label $y$ is produced as 
\begin{equation}
y = softmax(W \cdot \boldsymbol{c})
\end{equation}
where $W$ is a parameter. The term $\boldsymbol{c}$ is computed as a weighted sum of hidden vectors:
\begin{equation}
\boldsymbol{c} = \sum_{i=0}^{T}\alpha_{i}\boldsymbol{h}_i
\end{equation}
where $\alpha_{i}$ is the weight of $h_{i}$. The term $\boldsymbol{h}_{i}$ is the output of LSTM at the $i$-th word. The term $\alpha_{i}$ is computed as 
\begin{equation}
\alpha_i = \frac{\exp(e_i)}{\sum_{i=0}^{T}\exp(e_i)}
\end{equation}
where $e_i = f(\boldsymbol{h}_i, \boldsymbol{h}_T)$ is an alignment model.  We consider the last hidden state $\boldsymbol{h}_T$ as the context vector, which contains all information of an input sequence. The term $e_i$ evaluates the contribution of each word for sentiment classification. 

Our experimental results show that the proposed sentiment classifier achieves the accuracy of 89\% and 90\% on two datasets. With high classification accuracy, the attention weight produced by the classifier is considered to adequately capture the sentiment information of each word.

To extract non-emotional words based on continuous attention weights, we map attention weights to discrete values, 0 and 1. Since the discrete method is not the key part is this paper, we only use the following method for simplification.

We first calculate the averaged attention value in a sentence as
\begin{equation}
\bar{\alpha} = \frac{1}{T}\sum_{i=0}^{T}\alpha_i
\end{equation}
where $\bar{\alpha}$ is used as the threshold to distinguish non-emotional words from emotional words. The discrete attention weight is calculated  as
\begin{equation}
\hat{\alpha_i} = \left\{
\begin{aligned}
1, \hspace{2mm} \text{if} \hspace{2mm} \alpha_i \leq \bar{\alpha} \\
0, \hspace{2mm} \text{if} \hspace{2mm} \alpha_i > \bar{\alpha}
\end{aligned}
\right.
\end{equation}
where $\hat{\alpha_i}$ is treated as the identifier.

For pre-training the neutralization module, we build the training pair of input text $\boldsymbol{x}$ and a discrete attention weight sequence $\hat{ \boldsymbol{\alpha}}$. The cross entropy loss is computed as
\begin{equation}\label{newi}
L_{\theta} = -\sum_{i=1}^{T}P_{N_{\theta}}(\hat{\alpha _i}|x_i)
\end{equation}

\subsection{Emotionalization Module}\label{Emotionalization}
The emotionalization module $E_{\phi}$ is responsible for adding sentiment to the neutralized semantic content. In our work, we use a bi-decoder based encoder-decoder framework, which contains one encoder and two decoders. One decoder adds the positive sentiment and the other adds the negative sentiment. The input sentiment signal determines which decoder to use. Specifically, we use the seq2seq model ~\cite{DBLP:conf/nips/SutskeverVL14} for implementation. Both the encoder and decoder are LSTM networks. The encoder learns to compress the semantic content into a dense vector. The decoder learns to add sentiment based on the dense vector. Given the neutralized semantic content and the target sentiment, this module is responsible for producing an emotional sequence. 

For pre-training the emotionalization module, we first generate a neutralized input sequence $\hat{\boldsymbol{x}}$ by removing emotional words identified by the proposed sentiment classifier. Given the training pair of a neutralized sequence $\hat{\boldsymbol{x}}$ and an original sentence $\boldsymbol{x}$ with sentiment $s$, the cross entropy loss is computed as
\begin{equation}\label{em}
L_{\phi} = -\sum_{i=1}^{T}P_{E_{\phi}}(x_i|\hat{x}_i, s) 
\end{equation}
where a positive example goes through the positive decoder and a negative example goes through the negative decoder. 

We also explore a simpler method for pre-training the emotionalization module, which uses the product between a continuous vector $1-\boldsymbol{\alpha}$ and a word embedding sequence as the neutralized content where $\boldsymbol{\alpha}$ represents an attention weight sequence. Experimental results show that this method achieves much lower results than explicitly removing emotional words  based on discrete attention weights. Thus, we do not choose this method in our work.

\subsection{Cycled Reinforcement Learning}\label{dual}

Two modules are trained by the proposed cycled method. The neutralization module first neutralizes an emotional input to semantic content and then the emotionalization module is forced to reconstruct the original sentence based on the source sentiment and the semantic content. Therefore, the emotionalization module is taught to add sentiment to the semantic content in a supervised way. Because of the discrete choice of neutral words, the loss is no longer differentiable over the neutralization module. Therefore, we formulate it as a reinforcement learning problem and use policy gradient to train the neutralization module. The detailed training process is shown as follows.

We refer the neutralization module $N_{\theta}$ as the first agent and the emotionalization module $E_{\phi}$ as the second one. Given a sentence $\boldsymbol{x}$ associated with sentiment $s$, the term $\hat{\boldsymbol{x}}$ represents the middle neutralized context extracted by $\hat{\boldsymbol{\alpha}}$, which is generated by $P_{N_{\theta}}(\hat{ \boldsymbol{\alpha}} |\boldsymbol{x})$. 

In cycled training, the original sentence can be viewed as the supervision for training the second agent. Thus, the gradient for the second agent is
\begin{equation}\label{dual-learning-2}
\begin{split}
\nabla_{\phi}J(\phi) = \nabla_{\phi} \log(P_{E_{\phi}}(\boldsymbol{x} |\hat{\boldsymbol{x}}, s))
\end{split}
\end{equation}

We denote $\bar{\boldsymbol{x}}$ as the output generated by $P_{E_{\phi}}(\bar{\boldsymbol{x}} |\hat{\boldsymbol{x}}, s)$. We also denote $\boldsymbol{y}$ as the output  generated by $P_{E_{\phi}}(\boldsymbol{y }|\hat{\boldsymbol{x}}, \bar{s})$ where $\bar{s}$ represents the opposite sentiment.  Given $\bar{\boldsymbol{x}}$ and $\boldsymbol{y}$, we first calculate rewards for training the neutralized module, $R_{1}$ and $R_{2}$. The details of calculation process will be introduced in Section~\ref{reward-sec}. Then, we optimize parameters through policy gradient by maximizing the expected reward to train the neutralization module. It guides the neutralization module to identify non-emotional words better. In return, the improved neutralization module further enhances the emotionalization module.

According to the policy gradient theorem~\citep{DBLP:journals/ml/Williams92}, the gradient for the first agent is
\begin{equation}\label{dual-learning-1}
\begin{split}
\nabla_{\theta}J(\theta)= \mathbb{E}[R_{c} \cdot \nabla_{\theta}\log(P_{N_{\theta}}(\hat{\boldsymbol{\alpha}} |\boldsymbol{x}))]
\end{split}
\end{equation}
where $R_{c}$ is calculated as
\begin{equation}\label{combined_reward}
\begin{split}
R_{c} = R_{1}+R_{2}
\end{split}
\end{equation}

Based on Eq.~\ref{dual-learning-2} and Eq.~\ref{dual-learning-1}, we use the sampling approach to estimate the expected reward. This cycled process is repeated until converge.

\subsubsection{Reward}\label{reward-sec}

The reward consists of two parts, sentiment confidence and BLEU. Sentiment confidence evaluates whether the generated text matches the target sentiment. We use a pre-trained classifier to make the judgment. Specially, we use the proposed self-attention based sentiment classifier for implementation. The BLEU~\cite{DBLP:conf/acl/PapineniRWZ02} score is used to measure the content preservation performance. Considering that the reward should encourage the model to improve both metrics, we use the harmonic mean of sentiment confidence and BLEU as reward, which is formulated as 
\begin{equation}\label{reward}
R = (1+\beta^2) \frac{BLEU \cdot Confid}{ (\beta^2 \cdot BLEU) + Confid}
\end{equation}
where $\beta$ is a harmonic weight.

\section{Experiment}

In this section, we evaluate our method on two review datasets. We first introduce the datasets, the training details, the baselines, and the evaluation metrics. Then, we compare our approach with the state-of-the-art systems.  Finally, we show the experimental results and provide the detailed analysis of the key components.

\subsection{Unpaired Datasets}

We conduct experiments on two review datasets that contain user ratings associated with each review. Following previous work~\cite{Shen17crossalign}, we consider reviews with rating above three as positive reviews and reviews below three as negative reviews. The positive and negative reviews are not paired. Since our approach focuses on sentence-level sentiment-to-sentiment translation where sentiment annotations are provided at the document level, we process the two datasets with the following steps. First, following previous work~\cite{Shen17crossalign}, we filter out the reviews that exceed 20 words. Second, we construct text-sentiment pairs by extracting the first sentence in a review associated with its sentiment label, because the first sentence usually expresses the core idea. Finally, we train a sentiment classifier and filter out the text-sentiment pairs with the classifier confidence below 0.8. Specially, we use the proposed self-attention based sentiment classifier for implementation. The details of the processed datasets are introduced as follows. 

\textbf{Yelp Review Dataset (Yelp)}: This dataset is provided by Yelp Dataset Challenge.\footnote{\url{https://www.yelp.com/dataset/challenge}} The processed Yelp dataset contains  1.43M, 10K, and 5K pairs for training, validation, and testing, respectively. 

\textbf{Amazon Food Review Dataset (Amazon):} This dataset is provided by~\newcite{McAuley2013}. It consists of amounts of food reviews from Amazon.\footnote{\url{http://amazon.com}} The processed Amazon dataset contains 367K, 10K, and 5K pairs for training, validation, and testing, respectively.

\subsection{Training Details}

We tune hyper-parameters based on the performance on the validation sets. The self-attention based sentiment classifier is trained for 10 epochs on two datasets. We set $\beta$ for calculating reward to 0.5, hidden size to 256, embedding size to 128, vocabulary size to 50K, learning rate to 0.6, and batch size to 64. We use the Adagrad~\citep{DBLP:journals/jmlr/DuchiHS11} optimizer. All of the gradients are clipped when the norm exceeds 2. Before cycled training, the neutralization module and the emotionalization module are pre-trained for 1 and 4 epochs on the yelp dataset, for 3 and 5 epochs on the Amazon dataset.

\subsection{Baselines}

We compare our proposed method with the following state-of-the-art systems. 

\textbf{Cross-Alignment Auto-Encoder (CAAE)}: This method is proposed by~\citet{Shen17crossalign}. They propose a method that uses refined alignment of latent representations in hidden layers to perform style transfer. We treat this model as a baseline and adapt it by using the released code.

\textbf{Multi-Decoder with Adversarial Learning (MDAL)}: This method is proposed by~\citet{DBLP:journals/corr/abs-1711-06861}. They use a multi-decoder model with adversarial learning to separate style representations and content representations in hidden layers. We adapt this model by using the released code.

\subsection{Evaluation Metrics}

We conduct two evaluations in this work, including an automatic evaluation and a human evaluation. The details of evaluation metrics are shown as follows.

\subsubsection{Automatic Evaluation}

We quantitatively measure sentiment transformation by evaluating the accuracy of generating designated sentiment. For a fair comparison, we do not use the proposed sentiment classification model. Following previous work~\cite{Shen17crossalign,Hu17controlled}, we instead use a state-of-the-art sentiment classifier~\citep{DBLP:conf/clei/VieiraM17}, called TextCNN, to automatically evaluate the transferred sentiment accuracy. TextCNN achieves the accuracy of 89\% and 88\% on two datasets. Specifically, we generate sentences given sentiment $s$, and use the pre-trained sentiment classifier to assign sentiment labels to the generated sentences. The accuracy is calculated as the percentage of the predictions that match the sentiment $s$.

To evaluate the content preservation performance, we use the BLEU score~\cite{DBLP:conf/acl/PapineniRWZ02} between the transferred sentence and the source sentence as an evaluation metric. BLEU is a widely used metric for text generation tasks, such as machine translation, summarization, etc. The metric compares the automatically produced text with the reference text by computing overlapping lexical n-gram units.

To evaluate the overall performance, we use the geometric mean of ACC and BLEU as an evaluation metric. The G-score is one of the most commonly used ``single number" measures in Information Retrieval, Natural Language Processing, and Machine Learning.

\subsubsection{Human Evaluation}
While the quantitative evaluation provides indication of sentiment transfer quality, it can not evaluate the quality of transferred text accurately. Therefore, we also perform a human evaluation on the test set.  We randomly choose 200 items for the human evaluation.  Each item contains the transformed sentences generated by different systems given the same source sentence. The items are distributed to annotators who have no knowledge about which system the sentence is from. They are asked to score the transformed text in terms of sentiment and semantic similarity. Sentiment represents whether the sentiment of the source text is transferred correctly. Semantic similarity evaluates the context preservation performance. The score ranges from 1 to 10 (1 is very bad and 10 is very good).

\begin{table}[t]
\footnotesize
\centering
    \begin{tabular}{l|c|c|c}
    \hline
   
   Yelp & ACC & BLEU  &  G-score  \\\hline
    CAAE~\citep{Shen17crossalign} & 93.22  & 1.17 &10.44\\ 
    MDAL~\citep{DBLP:journals/corr/abs-1711-06861} & 85.65  & 1.64 & 11.85\\
   
    Proposed Method &80.00& 22.46 & \textbf{42.38} \\\hline \hline
    
     Amazon & ACC   & BLEU & G-score \\ \hline
    CAAE~\citep{Shen17crossalign} &84.19 & 0.56 &6.87\\ 
    MDAL~\citep{DBLP:journals/corr/abs-1711-06861} & 70.50& 0.27 & 4.36 \\ 
  
    Proposed Method & 70.37& 14.06 & \textbf{31.45}\\
    \hline

    \end{tabular}
    \caption{Automatic evaluations of the proposed method and baselines. ACC evaluates sentiment transformation. BLEU evaluates content preservation. G-score is the geometric mean of ACC and BLEU. } 
    \label{tab:state-div}
\end{table}

\subsection{Experimental Results}

Automatic evaluation results are shown in Table~\ref{tab:state-div}. ACC evaluates sentiment transformation. BLEU evaluates semantic content preservation. G-score represents the geometric mean of ACC and BLEU. CAAE and MDAL achieve much lower BLEU scores, 1.17 and 1.64 on the Yelp dataset, 0.56 and 0.27 on the Amazon dataset. The low BLEU scores indicate the worrying content preservation performance to some extent. Even with the desired sentiment, the irrelevant generated text leads to worse overall performance. In general, these two systems work more like sentiment-aware language models that generate text only based on the target sentiment and neglect the source input. The main reason is that these two systems attempt to separate emotional information from non-emotional content in a hidden layer, where all information is complicatedly mixed together. It is difficult to only  modify emotional information without any loss of non-emotional semantic content.


\begin{table}[t]
\footnotesize
\setlength{\tabcolsep}{3pt}
\centering
    \begin{tabular}{l|c|c|c}
    \hline
    Yelp & Sentiment & Semantic  & G-score\\ \hline
   CAAE~\citep{Shen17crossalign} & 7.67 & 3.87 & 5.45  \\
   MDAL~\citep{DBLP:journals/corr/abs-1711-06861} & 7.12& 3.68 & 5.12\\
  
   Proposed Method & 6.99 & 5.08 & \textbf{5.96} \\
   
    \hline \hline
    Amazon& Sentiment & Semantic  & G-score\\ \hline
   CAAE~\citep{Shen17crossalign} & 8.61 & 3.15 &  5.21\\
   MDAL~\citep{DBLP:journals/corr/abs-1711-06861} &7.93 & 3.22 & 5.05\\
  
   Proposed Method & 7.92 & 4.67 &  \textbf{6.08}\\ \hline
    
    \end{tabular}
    \caption{Human evaluations of the proposed method and baselines. \textsl{Sentiment} evaluates sentiment transformation. \textsl{Semantic} evaluates content preservation.  }
    \label{human}
    
\vspace{-1.0\baselineskip}
\end{table}

In comparison, our proposed method achieves the best overall performance on the two datasets, demonstrating the ability of learning knowledge from unpaired data. This result is attributed to the improved BLEU score. The BLEU score is largely improved from 1.64 to 22.46 and from 0.56 to 14.06 on the two datasets. The score improvements mainly come from the fact that we  separate emotional information from semantic content by explicitly filtering out emotional words. The extracted content is preserved and fed into the emotionalization module. Given the overall quality of transferred text as the reward, the neutralization module is taught to extract non-emotional semantic content better.  

Table ~\ref{human} shows the human evaluation results. It can be clearly seen that the proposed method obviously improves semantic preservation. The semantic score is increased from 3.87 to 5.08 on the Yelp dataset, and from 3.22 to 4.67 on the Amazon dataset. In general, our proposed model achieves the best overall performance. Furthermore, it also needs to be noticed that with the large improvement in  content preservation, the sentiment accuracy of the proposed method is lower than that of CAAE on the two datasets. It shows that simultaneously promoting sentiment transformation and content preservation remains to be studied further.

By comparing two evaluation results, we find that there is an agreement between the human evaluation and the automatic evaluation. It indicates the usefulness of automatic evaluation metrics. However, we also notice that the human evaluation has a smaller performance gap between the baselines and the proposed method than the automatic evaluation. It shows the limitation of automatic metrics for giving accurate results. For evaluating sentiment transformation, even with a high accuracy, the sentiment classifier sometimes generates noisy results, especially for those neutral sentences  (e.g., ``I ate a cheese sandwich'').  For evaluating content preservation, the BLEU score is computed based on the percentage of overlapping n-grams between the generated text and the reference text. However, the overlapping n-grams contain not only content words but also function words, bringing the noisy results. In general, accurate automatic evaluation metrics are expected in future work.

\begin{table}[t]
\footnotesize
\centering
    \begin{tabular}{p{0.9\linewidth}}
    \hline
       \textbf{Input}: \textsl{I would strongly advise against using this company.}\\
       \textbf{CAAE}:  \textsl{I love this place for a great experience here.}    \\  
      \textbf{MDAL}: \textsl{I have been a great place was great.}\\

    \textbf{Proposed Method}: \textsl{I would love using this company.}\\ \hline

     \textbf{Input}: \textsl{The service was nearly non-existent and extremely rude.}\\
      \textbf{CAAE}:  \textsl{The best place in the best area in vegas.}    \\  
      \textbf{MDAL}: \textsl{The food is very friendly and very good.
}\\

    \textbf{Proposed Method}: \textsl{The service was served and completely fresh.}\\ \hline

      \textbf{Input}: \textsl{Asked for the roast beef and mushroom sub, only received roast beef.}\\
      \textbf{CAAE}:  \textsl{We had a great experience with. }    \\ 
      \textbf{MDAL}: \textsl{This place for a great place for a great food and best.}\\

    \textbf{Proposed Method}: \textsl{Thanks for the beef and spring bbq.}\\ \hline

      \textbf{Input}: \textsl{Worst cleaning job ever!}\\
      \textbf{CAAE}:  \textsl{Great food and great service!}    \\  
      \textbf{MDAL}: \textsl{Great food, food!}\\

    \textbf{Proposed Method}: \textsl{Excellent outstanding job ever!}\\ \hline
    
      \textbf{Input}: \textsl{Most boring show I've ever been.}\\
      \textbf{CAAE}:  \textsl{Great place is the best place in town.}    \\ 
      \textbf{MDAL}: \textsl{Great place I've ever ever had.}\\

    \textbf{Proposed Method}: \textsl{Most amazing show I've ever been.}\\ \hline

    \textbf{Input}: \textsl{Place is very clean and the food is delicious.}\\
      \textbf{CAAE}:  \textsl{Don't go to this place.}    \\ 
      \textbf{MDAL}: \textsl{This place wasn't worth the worst place is horrible.}\\
    
    \textbf{Proposed Method}: \textsl{Place is very small and the food is terrible.}\\ 
    
    \hline

    \textbf{Input}: \textsl{Really satisfied with experience buying clothes.}\\
      \textbf{CAAE}:  \textsl{Don't go to this place.}    \\ 
      \textbf{MDAL}: \textsl{Do not impressed with this place.}\\
    
    \textbf{Proposed Method}: \textsl{Really bad experience.}\\

    \hline  
    \end{tabular}
    \caption{Examples generated by the proposed approach and baselines on the Yelp dataset. The two baselines change not only the polarity of examples, but also the semantic content. In comparison, our approach changes the sentiment of sentences with higher semantic similarity.}
    \label{samplecases}
\vspace{-1\baselineskip}
\end{table}

Table \ref{samplecases} presents the examples generated by different systems on the Yelp dataset. The two baselines change not only the polarity of examples, but also the semantic content. In comparison, our method precisely changes the sentiment of sentences (and paraphrases slightly to ensure fluency), while keeping the semantic content unchanged.

\begin{table}[t]
\footnotesize
\setlength{\tabcolsep}{1.5pt}
\centering
    \begin{tabular}{l|c|c|c}
    \hline
   Yelp &  ACC & BLEU &G-score \\\hline

    Emotionalization Module &  41.84 & 25.66& 32.77 \\ 
   
    + NM + Cycled RL & 85.71 & 1.08 & 9.62  \\
    + NM + Pre-training &  70.61& 17.02 & 34.66  \\
    + NM + Cycled RL + Pre-training& 80.00& 22.46& \textbf{42.38}  \\  
   \hline \hline
     Amazon & ACC & BLEU &G-score\\ \hline
   Emotionalization Module & 57.28 & 12.22 &26.46  \\ 
  
    + NM + Cycled RL & 64.16 & 8.03 & 22.69 \\ 
    + NM + Pre-training & 69.61& 11.16 & 27.87\\
    + NM + Cycled RL + Pre-training & 70.37& 14.06 & \textbf{31.45}\\
    \hline

    \end{tabular}
    \caption{Performance of key components in the  proposed approach. ``NM'' denotes the neutralization module. ``Cycled RL'' represents cycled reinforcement learning. }
    \label{module}
    \vspace{-1.5\baselineskip}
\end{table}

\subsection{Incremental Analysis}

In this section, we conduct a series of experiments to evaluate the contributions of our key components. The results are shown in Table~\ref{module}.  

We treat the emotionalization module as a baseline where the input is the original emotional sentence. The emotionalization module achieves the highest BLEU score but with much lower sentiment transformation accuracy. The encoding of the original sentiment leads to the emotional hidden vector that influences the decoding process and results in worse sentiment transformation performance. 

It can be seen that the method with all components achieves the best performance. First, we find that the method that only uses cycled reinforcement learning performs badly because it is hard to guide two randomly initialized modules to teach each other. Second, the pre-training method brings a slight improvement in overall performance. The G-score is improved from 32.77 to 34.66 and from 26.46 to 27.87 on the two datasets. The bottleneck of this method is the noisy attention weight because of the limited sentiment classification accuracy. Third, the method that combines cycled reinforcement learning and pre-training achieves the better performance than using one of them. Pre-training gives the two modules initial learning ability. Cycled training teaches the two modules to improve each other based on the feedback signals. Specially, the G-score is improved from 34.66 to 42.38 and from 27.87 to 31.45 on the two datasets. Finally, by comparing the methods with and without the neutralization module, we find that the neutralization mechanism improves a lot in sentiment transformation with a slight reduction on content preservation. It proves the effectiveness of explicitly separating sentiment information from semantic content.

Furthermore, to analyze the neutralization ability in the proposed method, we randomly sample several examples, as shown in Table~\ref{anasamle}. It can be clearly seen that emotional words are removed accurately almost without loss of non-emotional information.

\subsection{Error Analysis}

Although the proposed method outperforms the state-of-the-art systems, we also observe several failure cases, such as sentiment-conflicted sentences (e.g., ``Outstanding and bad service''), neutral sentences (e.g., ``Our first time here''). Sentiment-conflicted sentences indicate that the original sentiment is not removed completely. This problem occurs when the input contains emotional words that are unseen in the training data, or the sentiment is implicitly expressed. Handling complex sentiment expressions  is an important problem for future work. Neutral sentences demonstrate that the decoder sometimes fails in adding the target sentiment and only generates text based on the semantic content. A better sentiment-aware decoder is expected to be explored in future work.

\begin{table}[t]
\footnotesize
\centering
    \begin{tabular}{p{0.95\linewidth}}
    \hline
         \textsl{ Michael is absolutely \textcolor{red}{wonderful}.}\\ 
         \textsl{ I would strongly advise} \textcolor{red}{against} \textsl{using this company.} \\ 
          \textsl{\textcolor{red}{Horrible} experience!}\\
          \textsl{\textcolor{red}{Worst cleaning} job ever!}\\ 
           \textsl{Most \textcolor{red}{boring} show i 've ever been.}\\ 
           \textsl{Hainan chicken was really \textcolor{red}{good}.}\\ 
            \textsl{I really don't understand all the \textcolor{red}{negative reviews} for this dentist.}\\
            \textsl{Smells \textcolor{red}{so weird} in there.}
            
            \textsl{The service was nearly \textcolor{red}{non-existent} and extremely \textcolor{red}{ rude}.}\\

    \hline  
    \end{tabular}
    \caption{Analysis of the neutralization module. Words in red are removed by the neutralization module.   }
    \label{anasamle}
\vspace{-1.5\baselineskip}
\end{table}

\section{Conclusions and Future Work}

In this paper, we focus on unpaired sentiment-to-sentiment translation and propose a cycled reinforcement learning approach that enables training in the absence of parallel training data. We conduct experiments on two review datasets. Experimental results show that our method substantially outperforms the state-of-the-art systems, especially in terms of semantic preservation. For future work, we would like to explore a fine-grained version of sentiment-to-sentiment translation that not only reverses sentiment, but also changes the strength of sentiment. 

\section*{Acknowledgements}

This work was supported in part by National Natural Science Foundation of China (No. 61673028), National High Technology Research and Development Program of China (863 Program, No. 2015AA015404), and the National Thousand Young Talents Program. Xu Sun is the corresponding author of this paper.
\nocite{dong2017learning,zang2017towards,DBLP:journals/corr/abs-1803-01557,DBLP:journals/corr/abs-1803-01465,DBLP:journals/corr/abs-1802-01345,DBLP:journals/corr/abs-1802-01812}


\bibliography{acl2018}

\begin{thebibliography}{28}
\expandafter\ifx\csname natexlab\endcsname\relax\def\natexlab#1{#1}\fi

\bibitem[{Chen et~al.(2017)Chen, Xu, He, and
  Wang}]{DBLP:journals/eswa/ChenXHW17}
Tao Chen, Ruifeng Xu, Yulan He, and Xuan Wang. 2017.
\newblock Improving sentiment analysis via sentence type classification using
  bilstm-crf and {CNN}.
\newblock \emph{Expert Syst. Appl.}, 72:221--230.

\bibitem[{Dong et~al.(2017)Dong, Huang, Wei, Lapata, Zhou, and
  Xu}]{dong2017learning}
Li~Dong, Shaohan Huang, Furu Wei, Mirella Lapata, Ming Zhou, and Ke~Xu. 2017.
\newblock Learning to generate product reviews from attributes.
\newblock In \emph{Proceedings of the 15th Conference of the European Chapter
  of the Association for Computational Linguistics: Volume 1, Long Papers},
  volume~1, pages 623--632.

\bibitem[{Duchi et~al.(2011)Duchi, Hazan, and
  Singer}]{DBLP:journals/jmlr/DuchiHS11}
John~C. Duchi, Elad Hazan, and Yoram Singer. 2011.
\newblock Adaptive subgradient methods for online learning and stochastic
  optimization.
\newblock \emph{Journal of Machine Learning Research}, 12:2121--2159.

\bibitem[{Fu et~al.(2018)Fu, Tan, Peng, Zhao, and
  Yan}]{DBLP:journals/corr/abs-1711-06861}
Zhenxin Fu, Xiaoye Tan, Nanyun Peng, Dongyan Zhao, and Rui Yan. 2018.
\newblock Style transfer in text: Exploration and evaluation.
\newblock In \emph{AAAI 2018}.

\bibitem[{Gatys et~al.(2016)Gatys, Ecker, and
  Bethge}]{DBLP:conf/cvpr/GatysEB16}
Leon~A. Gatys, Alexander~S. Ecker, and Matthias Bethge. 2016.
\newblock Image style transfer using convolutional neural networks.
\newblock In \emph{2016 {IEEE} Conference on Computer Vision and Pattern
  Recognition, {CVPR} 2016, Las Vegas, NV, USA, June 27-30, 2016}, pages
  2414--2423.

\bibitem[{He et~al.(2016)He, Xia, Qin, Wang, Yu, Liu, and
  Ma}]{DBLP:conf/nips/HeXQWYLM16}
Di~He, Yingce Xia, Tao Qin, Liwei Wang, Nenghai Yu, Tie{-}Yan Liu, and
  Wei{-}Ying Ma. 2016.
\newblock Dual learning for machine translation.
\newblock In \emph{Advances in Neural Information Processing Systems 29: Annual
  Conference on Neural Information Processing Systems 2016, December 5-10,
  2016, Barcelona, Spain}, pages 820--828.

\bibitem[{Hu et~al.(2017)Hu, Yang, Liang, Salakhutdinov, and
  Xing}]{Hu17controlled}
Zhiting Hu, Zichao Yang, Xiaodan Liang, Ruslan Salakhutdinov, and Eric~P. Xing.
  2017.
\newblock Controllable text generation.
\newblock In \emph{ICML 2017}.

\bibitem[{Johnson et~al.(2016)Johnson, Alahi, and
  Fei-Fei}]{DBLP:conf/eccv/JohnsonAF16}
Justin Johnson, Alexandre Alahi, and Li~Fei-Fei. 2016.
\newblock Perceptual losses for real-time style transfer and super-resolution.
\newblock In \emph{ECCV, 2016}, pages 694--711.

\bibitem[{Li et~al.(2017)Li, Wang, Liu, and Hou}]{DBLP:conf/ijcai/li}
Yanghao Li, Naiyan Wang, Jiaying Liu, and Xiaodi Hou. 2017.
\newblock Demystifying neural style transfer.
\newblock In \emph{Proceedings of the Twenty-Sixth International Joint
  Conference on Artificial Intelligence, {IJCAI} 2017, Melbourne, Australia,
  August 19-25, 2017}, pages 2230--2236.

\bibitem[{Liao et~al.(2017)Liao, Yao, Yuan, Hua, and
  Kang}]{DBLP:journals/tog/LiaoYYHK17}
Jing Liao, Yuan Yao, Lu~Yuan, Gang Hua, and Sing~Bing Kang. 2017.
\newblock Visual attribute transfer through deep image analogy.
\newblock \emph{{ACM} Trans. Graph.}, 36(4):120:1--120:15.

\bibitem[{Lin et~al.(2018)Lin, Ma, Su, and
  Sun}]{DBLP:journals/corr/abs-1802-01812}
Junyang Lin, Shuming Ma, Qi~Su, and Xu~Sun. 2018.
\newblock Decoding-history-based adaptive control of attention for neural
  machine translation.
\newblock \emph{CoRR}, abs/1802.01812.

\bibitem[{Ma et~al.(2017)Ma, Li, Zhang, Wang, and
  Sun}]{DBLP:conf/ijcnlp/MaLZWS17}
Dehong Ma, Sujian Li, Xiaodong Zhang, Houfeng Wang, and Xu~Sun. 2017.
\newblock Cascading multiway attentions for document-level sentiment
  classification.
\newblock In \emph{Proceedings of the Eighth International Joint Conference on
  Natural Language Processing, {IJCNLP} 2017, Taipei, Taiwan, November 27 -
  December 1, 2017 - Volume 1: Long Papers}, pages 634--643.

\bibitem[{Ma et~al.(2018{\natexlab{a}})Ma, Sun, Li, Li, Li, and
  Ren}]{DBLP:journals/corr/abs-1803-01465}
Shuming Ma, Xu~Sun, Wei Li, Sujian Li, Wenjie Li, and Xuancheng Ren.
  2018{\natexlab{a}}.
\newblock Query and output: Generating words by querying distributed word
  representations for paraphrase generation.
\newblock \emph{CoRR}, abs/1803.01465.

\bibitem[{Ma et~al.(2018{\natexlab{b}})Ma, Sun, Lin, and Ren}]{shumingmaijcai}
Shuming Ma, Xu~Sun, Junyang Lin, and Xuancheng Ren. 2018{\natexlab{b}}.
\newblock A hierarchical end-to-end model for jointly improving text
  summarization and sentiment classification.
\newblock \emph{CoRR}, abs/1805.01089.

\bibitem[{McAuley and Leskovec(2013)}]{McAuley2013}
Julian~John McAuley and Jure Leskovec. 2013.
\newblock From amateurs to connoisseurs: modeling the evolution of user
  expertise through online reviews.
\newblock In \emph{22nd International World Wide Web Conference, {WWW} '13, Rio
  de Janeiro, Brazil, May 13-17, 2013}, pages 897--908.

\bibitem[{Papineni et~al.(2002)Papineni, Roukos, Ward, and
  Zhu}]{DBLP:conf/acl/PapineniRWZ02}
Kishore Papineni, Salim Roukos, Todd Ward, and Wei{-}Jing Zhu. 2002.
\newblock Bleu: a method for automatic evaluation of machine translation.
\newblock In \emph{Proceedings of the 40th Annual Meeting of the Association
  for Computational Linguistics, July 6-12, 2002, Philadelphia, PA, {USA.}},
  pages 311--318.

\bibitem[{Pontiki et~al.(2015)Pontiki, Galanis, Papageorgiou, Manandhar, and
  Androutsopoulos}]{DBLP:conf/semeval/PontikiGPMA15}
Maria Pontiki, Dimitris Galanis, Haris Papageorgiou, Suresh Manandhar, and Ion
  Androutsopoulos. 2015.
\newblock Semeval-2015 task 12: Aspect based sentiment analysis.
\newblock In \emph{Proceedings of the 9th International Workshop on Semantic
  Evaluation, SemEval@NAACL-HLT 2015, Denver, Colorado, USA, June 4-5, 2015},
  pages 486--495.

\bibitem[{Rosenthal et~al.(2017)Rosenthal, Farra, and
  Nakov}]{DBLP:conf/semeval/RosenthalFN17}
Sara Rosenthal, Noura Farra, and Preslav Nakov. 2017.
\newblock Semeval-2017 task 4: Sentiment analysis in twitter.
\newblock In \emph{Proceedings of the 11th International Workshop on Semantic
  Evaluation, SemEval@ACL 2017, Vancouver, Canada, August 3-4, 2017}, pages
  502--518.

\bibitem[{Shen et~al.(2017)Shen, Lei, Barzilay, and
  Jaakkola}]{Shen17crossalign}
Tianxiao Shen, Tao Lei, Regina Barzilay, and Tommi~S. Jaakkola. 2017.
\newblock Style transfer from non-parallel text by cross-alignment.
\newblock In \emph{NIPS 2017}.

\bibitem[{Socher et~al.(2011)Socher, Lin, Ng, and
  Manning}]{DBLP:conf/icml/SocherLNM11}
Richard Socher, Cliff~Chiung{-}Yu Lin, Andrew~Y. Ng, and Christopher~D.
  Manning. 2011.
\newblock Parsing natural scenes and natural language with recursive neural
  networks.
\newblock In \emph{Proceedings of the 28th International Conference on Machine
  Learning, {ICML} 2011, Bellevue, Washington, USA, June 28 - July 2, 2011},
  pages 129--136.

\bibitem[{Sutskever et~al.(2014)Sutskever, Vinyals, and
  Le}]{DBLP:conf/nips/SutskeverVL14}
Ilya Sutskever, Oriol Vinyals, and Quoc~V. Le. 2014.
\newblock Sequence to sequence learning with neural networks.
\newblock In \emph{Advances in Neural Information Processing Systems 27: Annual
  Conference on Neural Information Processing Systems 2014, December 8-13 2014,
  Montreal, Quebec, Canada}, pages 3104--3112.

\bibitem[{Tu et~al.(2017)Tu, Liu, Shang, Liu, and Li}]{DBLP:conf/aaai/TuLSLL17}
Zhaopeng Tu, Yang Liu, Lifeng Shang, Xiaohua Liu, and Hang Li. 2017.
\newblock Neural machine translation with reconstruction.
\newblock In \emph{Proceedings of the Thirty-First {AAAI} Conference on
  Artificial Intelligence, February 4-9, 2017, San Francisco, California,
  {USA.}}, pages 3097--3103.

\bibitem[{Vieira and Moura(2017)}]{DBLP:conf/clei/VieiraM17}
Joao Paulo~Albuquerque Vieira and Raimundo~Santos Moura. 2017.
\newblock An analysis of convolutional neural networks for sentence
  classification.
\newblock In \emph{{XLIII} 2017}, pages 1--5.

\bibitem[{Williams(1992)}]{DBLP:journals/ml/Williams92}
Ronald~J. Williams. 1992.
\newblock Simple statistical gradient-following algorithms for connectionist
  reinforcement learning.
\newblock \emph{Machine Learning}, 8:229--256.

\bibitem[{Xu et~al.(2018)Xu, Sun, Ren, Lin, Wei, and
  Li}]{DBLP:journals/corr/abs-1802-01345}
Jingjing Xu, Xu~Sun, Xuancheng Ren, Junyang Lin, Bingzhen Wei, and Wei Li.
  2018.
\newblock {DP-GAN:} diversity-promoting generative adversarial network for
  generating informative and diversified text.
\newblock \emph{CoRR}, abs/1802.01345.

\bibitem[{Zang and Wan(2017)}]{zang2017towards}
Hongyu Zang and Xiaojun Wan. 2017.
\newblock Towards automatic generation of product reviews from aspect-sentiment
  scores.
\newblock In \emph{Proceedings of the 10th International Conference on Natural
  Language Generation}, pages 168--177.

\bibitem[{Zhang et~al.(2018)Zhang, Li, and
  Sun}]{DBLP:journals/corr/abs-1803-01557}
Zhiyuan Zhang, Wei Li, and Xu~Sun. 2018.
\newblock Automatic transferring between ancient chinese and contemporary
  chinese.
\newblock \emph{CoRR}, abs/1803.01557.

\bibitem[{Zhu et~al.(2017)Zhu, Park, Isola, and
  Efros}]{DBLP:conf/iccv/ZhuPIE17}
Jun{-}Yan Zhu, Taesung Park, Phillip Isola, and Alexei~A. Efros. 2017.
\newblock Unpaired image-to-image translation using cycle-consistent
  adversarial networks.
\newblock In \emph{{IEEE} International Conference on Computer Vision, {ICCV}
  2017, Venice, Italy, October 22-29, 2017}, pages 2242--2251.

\end{thebibliography}
\bibliographystyle{acl_natbib}

\end{document}